\documentclass[runningheads]{llncs}

\usepackage{graphicx}
\usepackage{booktabs}
\usepackage{multirow}
\usepackage{comment}
\usepackage{url}
\usepackage{color, colortbl}
\usepackage[pagebackref,breaklinks,colorlinks]{hyperref}

\usepackage{amsmath}
\usepackage{amssymb}
\usepackage{bm}
\usepackage{caption}   

\definecolor{Gray}{gray}{0.95}
\newcolumntype{g}{>{\columncolor{Gray}}c}
\usepackage{here}
\usepackage{pifont}
\newcommand{\cmark}{\ding{51}} 
\newcommand{\xmark}{\ding{55}} 

\usepackage[T1]{fontenc}
\usepackage{graphicx,verbatim}
\begin{document}
\title{EgoSurgery-HTS: A Dataset for Egocentric Hand-Tool Segmentation in Open Surgery Videos}

\author{Nathan Darjana\inst{1*} \and
Ryo Fujii\inst{1*\dagger}\orcidID{0000-0002-9115-8414} \and
\\ Hideo Saito\inst{1}\orcidID{0000-0002-2421-9862} \and Hiroki Kajita\inst{2}}
\authorrunning{N. Darjana et al.}
%
\institute{Keio University, Yokohama, Kanagawa, Japan \\ \email{\{nathan.darjana, ryo.fujii0112, hs\}@keio.jp} \and
Keio University School of Medicine, Shinjuku, Tokyo, Japan\\
\email{\{jmrbx767\}@keio.jp}
}


\maketitle              
\def\thefootnote{*}\footnotetext{Corresponding authors.}\def\thefootnote{\arabic{footnote}}
\def\thefootnote{	$\dagger$}\footnotetext{Project lead.}\def\thefootnote{\arabic{footnote}}
\begin{abstract}
Egocentric open-surgery videos capture rich, fine-grained details essential for accurately modeling surgical procedures and human behavior in the operating room. A detailed, pixel-level understanding of hands and surgical tools is crucial for interpreting a surgeon’s actions and intentions. We introduce EgoSurgery-HTS, a new dataset with pixel-wise annotations and a benchmark suite for segmenting surgical tools, hands, and interacting tools in egocentric open-surgery videos. Specifically, we provide a labeled dataset for (1) tool instance segmentation of 14 distinct surgical tools, (2) hand instance segmentation, and (3) hand-tool segmentation to label hands and the tools they manipulate. Using EgoSurgery-HTS, we conduct extensive evaluations of state-of-the-art segmentation methods and demonstrate significant improvements in the accuracy of hand and hand-tool segmentation in egocentric open-surgery videos compared to existing datasets. The dataset will be released at \href{https://github.com/Fujiry0/EgoSurgery}{project page}.


\end{abstract}
\section{Introduction}
The automated analysis of egocentric open-surgery videos 
plays an important role in applications such as real-time surgical assistance, skill assessment, and medical procedure evaluation~\cite{liu2021towards}.

By providing fine-grained procedural insights, automated analysis has the potential to enhance surgical precision, reduce operative duration, and improve patient outcomes. A key aspect of this analysis is surgical scene segmentation, which enables per-pixel understanding of the operative field. In particular, the segmentation of hands and surgical tools is essential for interpreting a surgeon’s actions and intent, facilitating workflow optimization and AI-assisted surgery.

\begin{table*}[tb]
\caption{Comparison of EgoSurgery-HTS with existing surgical segmentation datasets.}
\begin{center}
\resizebox{\textwidth}{!}{
\begin{tabular}{lcccccc}
\toprule
Dataset  & Surgery Type & Frames & Tool Types & Hand Types & Hand-Tool Interaction \\
\midrule
Endovis2015~\cite{bodenstedt2018comparative}   &  \multirow{7}{*}{MIS}  & 10k  & 3  & -  & \xmark \\  
Endovis2017~\cite{allan20192017}   &  & 2.4k  & 10  & -  & \xmark \\ 
Endovis2018~\cite{allan20202018}   &  & 2.4k   & 10  & -  & \xmark \\ 
CholecSeg8k~\cite{hong2020cholecseg8k}   &  & 8k  & 12  & -  & \xmark \\ 
AutoLaparo~\cite{wang2022autolapar}   &  & 1.8k  & 7   & -  & \xmark \\ 
ROBUSTMIS2019~\cite{ross2021robust}  &  & 10k & 2   & -  & \xmark \\ 
SAR-RARP50~\cite{psychogyios2023sar} &  & 10k  & 10  & -  & \xmark \\ \hline
\rowcolor{Gray} EgoSurgery-HTS (Ours) & Open Surgery & 15.4K  & 14  & 4  & \cmark  \\
\bottomrule
\end{tabular}}
\end{center}
\label{tab:survey}
\end{table*}

Surgical tool segmentation, which involves the precise identification and delineation of surgical instruments, has gained significant attention, particularly in minimally invasive surgeries (MIS), leading to the development of various advanced approaches~\cite{baby2023fromforks,gonzalez2020ISINet,Islam2020apmtl,ni2020pyramid,shvets2018automatic,yue_surgicalsam,zhao2020learning}. A key factor driving this progress is the availability of large, well-annotated datasets~\cite{allan20202018,allan20192017,bodenstedt2018comparative,hong2020cholecseg8k,psychogyios2023sar,ross2021robust,wang2022autolapar}. While these advancements have greatly benefited MIS, tool segmentation in open surgery remains relatively underexplored As shown in Table~\ref{tab:survey}, existing datasets primarily focus on MIS, capturing specific perspectives, tools, environments, and procedural types that differ from those in open surgery. As a result, they fail to address the unique challenges inherent to open surgical procedures. These challenges include the frequent manipulation of multiple tools by multiple individuals from various angles, along with variations in lighting and camera perspectives. The absence of large-scale datasets specifically designed for open surgery significantly impedes progress in achieving accurate tool segmentation in surgical videos.

Hand segmentation is crucial for egocentric open surgery video understanding, as a surgeon’s hands are central to nearly every frame, playing a key role in instrument manipulation and surgical workflow analysis. While large-scale hand segmentation datasets exist for daily activities~\cite{Bambach2015egohand,Khan2018handsegwild}, they do not generalize well to surgery due to differences in appearance and motion~\cite{Zhang2021AMIA}. Despite its importance, research on hand segmentation in open surgery is limited, highlighting the need for a large-scale, domain-specific dataset.

Egocentric hand-object segmentation (HOS)~\cite{Darkhalil2022VISOR,Narasimhaswamy2024HOISTFormer,zhang2022fine}, which focuses on segmenting hands and interacting objects in egocentric videos, is essential for understanding the viewer’s behavior and intentions. Unlike minimally invasive surgery (MIS), where instruments are manipulated through small incisions, open surgery requires direct interaction with tissues and tools, making egocentric hand-object segmentation crucial for comprehending surgical procedures from a first-person perspective. However, existing datasets~\cite{Darkhalil2022VISOR,Narasimhaswamy2024HOISTFormer,zhang2022fine} primarily focus on daily activities~\cite{Tang2017Action,Damen2021epic}, limiting their applicability to surgical scenarios. To advance the automated analysis of egocentric open surgery videos, the development of annotated hand-object segmentation datasets tailored to the surgical domain is indispensable.

Building upon the EgoSurgery dataset~\cite{fujii2022surgicaltool,fujii2024egosurgeryphase,fujii2024egosurgerytool} and leveraging its comprehensive annotation suite for open surgery video understanding, we introduce EgoSurgery-HTS, a novel and detailed dataset designed to facilitate pixel-level analysis of open surgery scenes. Firstly, EgoSurgery-HTS provides tool instance segmentation annotations, offering fine-grained segmentation masks across $14$ distinct types of surgical tools. Secondly, it includes segmentation annotations for four types of hand instances. Finally, EgoSurgery-HTS features hand-object segmentation annotations, providing fine-grained per-pixel labels for hands and interacting objects. Using the proposed EgoSurgery-HTS dataset, we conduct a systematic study on mainstream segmentation baselines. Furthermore, with this new dataset, we significantly improve hand and hand-object segmentation performance compared to previous datasets in the open surgery domain, demonstrating the value and impact of EgoSurgery-HTS in advancing open surgery video analysis.

Our main contributions are summarized as follows: 1) We introduce EgoSurgery-HTS, a comprehensive dataset tailored for fine-grained understanding of egocentric open surgery. It includes detailed annotations for tool instance segmentation, hand instance segmentation, and hand-object segmentation, enabling advanced analysis of complex surgical scenes. 2) We conduct extensive evaluations of state-of-the-art instance segmentation methods for each task on EgoSurgery-HTS, and discuss their strengths and weaknesses. 3) Models trained on EgoSurgery-HTS demonstrate superior performance in hand and hand-object segmentation, significantly outperforming models trained on pre-existing datasets.

\section{Dataset}
\begin{figure*}[tb]
\centering
\includegraphics[width=\textwidth, trim=0 0 12cm 0, clip]{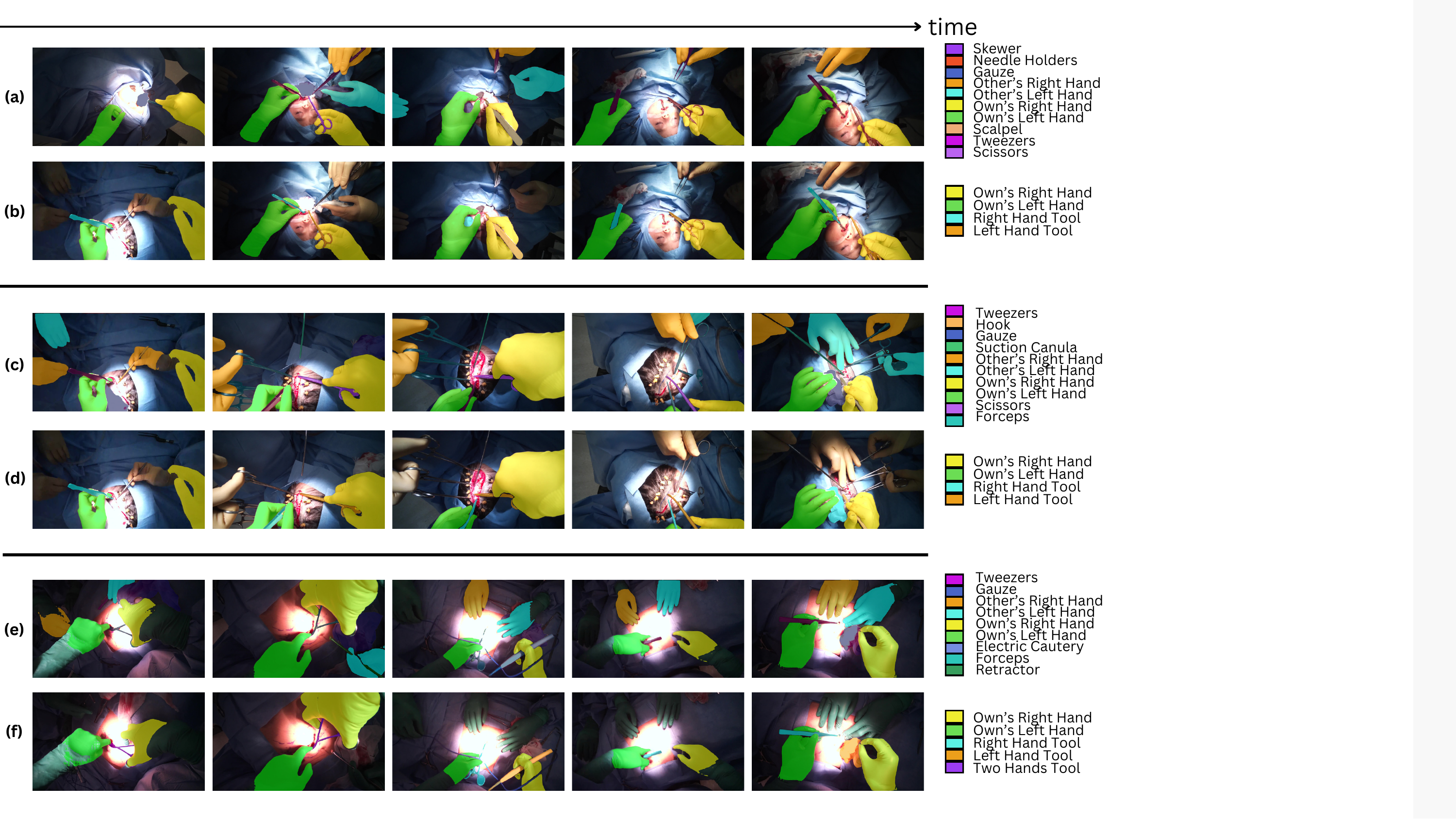}
\caption{Overview of the sparse segmentation from 2 different videos according to the different tasks. \textbf{(a)}-\textbf{(c)}-\textbf{(e)} Overview of every tool and hand instance segmentation task. \textbf{(b)}-\textbf{(d)}-\textbf{(f)} Overview of Hand-Object Segmentation task.  }
\label{dataset}
\end{figure*}

\subsection{Dataset Source and Annotations}
The EgoSurgery dataset~\cite{fujii2022surgicaltool,fujii2024egosurgeryphase,fujii2024egosurgerytool} comprises 21 videos spanning 10 distinct surgical procedures, with a total duration of 15 hours, performed by 8 surgeons. EgoSurgery provides over 27K frames with phase annotations and 15K frames with bounding box annotations for surgical tools and hands. However, EgoSurgery lacks per-pixel segmentation labels for surgical tools, hands, and their interactions. Therefore, we introduce EgoSurgery-HTS, an extension of EgoSurgery that includes additional annotations for surgical tool segmentation, hand segmentation, and hand-tool segmentation on a subset of the existing dataset. These comprehensive annotations establish EgoSurgery as the only available dataset enabling multi-task learning for phase recognition, surgical tool detection, hand detection, surgical tool segmentation, hand segmentation, and hand-tool segmentation.

\noindent \textbf{Annotation Process}: Inspired by SAMRS~\cite{wang2023samrs}, which leverages SAM~\cite{Kirillov_2023_ICCV} and existing remote sensing object detection datasets to construct a large-scale remote sensing segmentation dataset, we apply SAM to EgoSurgery, using tool and hand bounding box annotations to generate segmentation labels. For hand-tool segmentation annotations, interacting tool annotations are determined by selecting the tool with the highest IoU score relative to the hand segmentation. For each image in the dataset, we obtain segmentation masks for 14 types of surgical tools, along with per-pixel hand mask annotations where applicable. These hand-related masks include:
(a) own left hand, (b) own right hand, (c) other left hand, and (d) other right hand. Additionally, we provide hand-tool segmentation masks, categorized as:
(a) left-hand, (b) right-hand, (c) left-hand object, (d) right-hand object, and (e) two-hand object. All generated annotations undergo manual review and correction to ensure accuracy.

\subsection{Dataset Statistics}
\begin{figure*}[tb]
\centering
\includegraphics[height=0.18\textheight, width=\textwidth, trim = 1cm 18.5cm 3.5cm 0, clip]{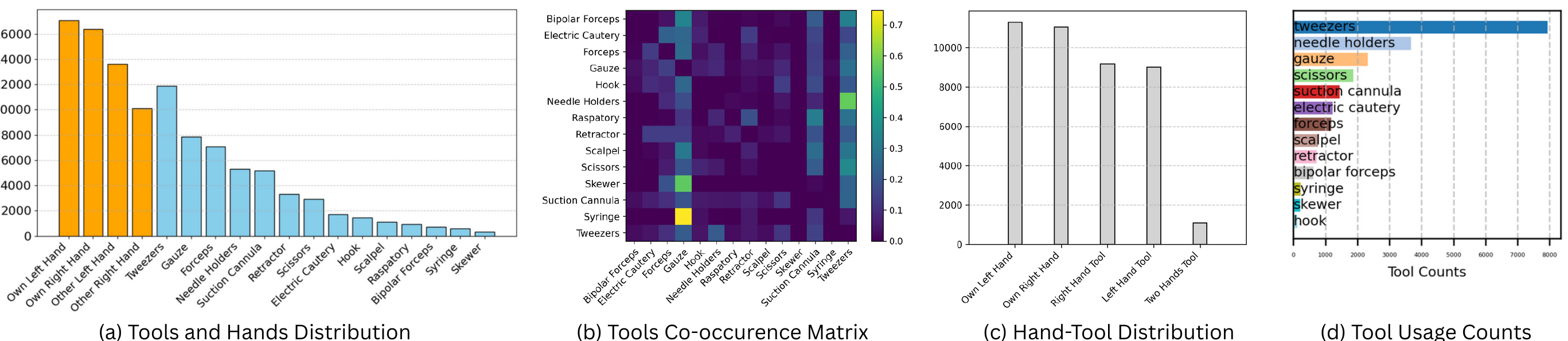}
\caption{Dataset Statistics \textbf{(a)} Distribution of different hands and surgical tools instances. \textbf{(b)} Frame level co-occurrence matrix tools in the dataset.  \textbf{(c)} Distribution of hands and its associated tools. \textbf{(d)} Tool usage counts based on manual handling. }
\label{fig:distribution}
\end{figure*}

The EgoSurgery-HTS dataset consists of 19,496 high-quality images, annotated with 50,383 tools, 57,173 hands, and 41,605 hand-tool segmentations. Fig.~\ref{fig:distribution} (a) illustrates a pronounced class imbalance among the different surgical tools. To further analyze tool co-occurrence patterns, we introduce a co-occurrence matrix in Fig.~\ref{fig:distribution} (b), which shows the probability of finding one tool given the presence of another. Notably, certain tools frequently appear together, such as syringes and gauze, or needle holders and tweezers. This highlights strong tool pairings like Syringe/Gauze and Needle Holder/Tweezers. Fig.~\ref{fig:distribution} (c) focuses on hand-tool segmentation, showcasing the distribution of hands and their associated tools. Additionally, it provides insights into the frequency of tool usage within the dataset. Tweezers and needle holders are used significantly more often than other surgical tools, indicating an uneven distribution of tool usage across surgical operations.

\section{Experiments and Benchmarking Methods}
\subsection{Experimental setups}
We evaluate four popular object detectors—Mask R-CNN~\cite{he2017maskrcnn}, QueryInst~\cite{Fang_2021_ICCV}, Mask2Former~\cite{cheng2021mask2former} and SOLOv2\cite{wang2020solov2}—on our dataset. The implementations are based on the MMDetection~\cite{mmdetection}, and we finetune models with pre-trained on MS-COCO~\cite{Lin2014coco}. To ensure a fair comparison, we select model backbones with a similar number of parameters. Following the original EgoSurgery, we adopt a video-split method for model training, considering the domain variations across videos. This approach ensures robust training by preventing models from overfitting to the specific surgical type.

\begin{table}[tb]
\caption{Experimental results ($\%$) of four state-of-the-art models on three different tasks: tool instance segmentation, hand instance segmentation, and hand-tool segmentation.}
\begin{center}
\resizebox{\textwidth}{!}{
\begin{tabular}{cccccccccc}
\toprule
 \multirow{2}{*}{Methods} & \multirow{2}{*}{Backbone}& \multicolumn{2}{c}{Tool}  &  \multicolumn{2}{c}{Hand} & \multicolumn{2}{c}{Hand-Tool}\\
  \cmidrule(l{2pt}r{3pt}){3-4}   \cmidrule(l{2pt}r{3pt}){5-6} \cmidrule(l{2pt}r{3pt}){7-8} 
 &   & mAP$^{\text{box}}$ & mAP$^{\text{mask}}$ & mAP$^{\text{box}}$ & mAP$^{\text{mask}}$   & mAP$^{\text{box}}$ & mAP$^{\text{mask}}$  \\

\hline
Mask R-CNN~\cite{he2017maskrcnn} & ResNet-50~\cite{he2016deep}  & 36.7 & 29.1 & \textbf{63.8} & \textbf{61.9} & 45.3 & 44.5 \\
QueryInst~\cite{Fang_2021_ICCV} &  ResNet-50~\cite{he2016deep}  & \textbf{47.3} & 36.7 & 54.0 & 50.7 & \textbf{55.2} & 49.4 \\
Mask2Former~\cite{cheng2021mask2former} &  ResNet-50~\cite{he2016deep} & 39.2 & \textbf{40.9} & 50.2 &52.5 & 54.7 & \textbf{56.6} \\SOLOv2~\cite{wang2020solov2} &  ResNet-50~\cite{he2016deep} & - & 37.0  &- & 53.8 & - & 50.7 \\
\bottomrule
\end{tabular}}
\end{center}
\label{tab:benchmark}
\end{table}

\subsection{Evaluation Metrics}
The experimental results are reported using the standard COCO metrics, where the average precision (AP) is computed as the mean intersection over union (IoU) across 10 thresholds ranging from 0.5 to 0.95 (at intervals of 0.05) for both bounding boxes (box AP) and segmentation masks (mask AP).

\subsection{Quantitative results}

\begin{figure*}[tb]
\centering
\includegraphics[width=1\textwidth]{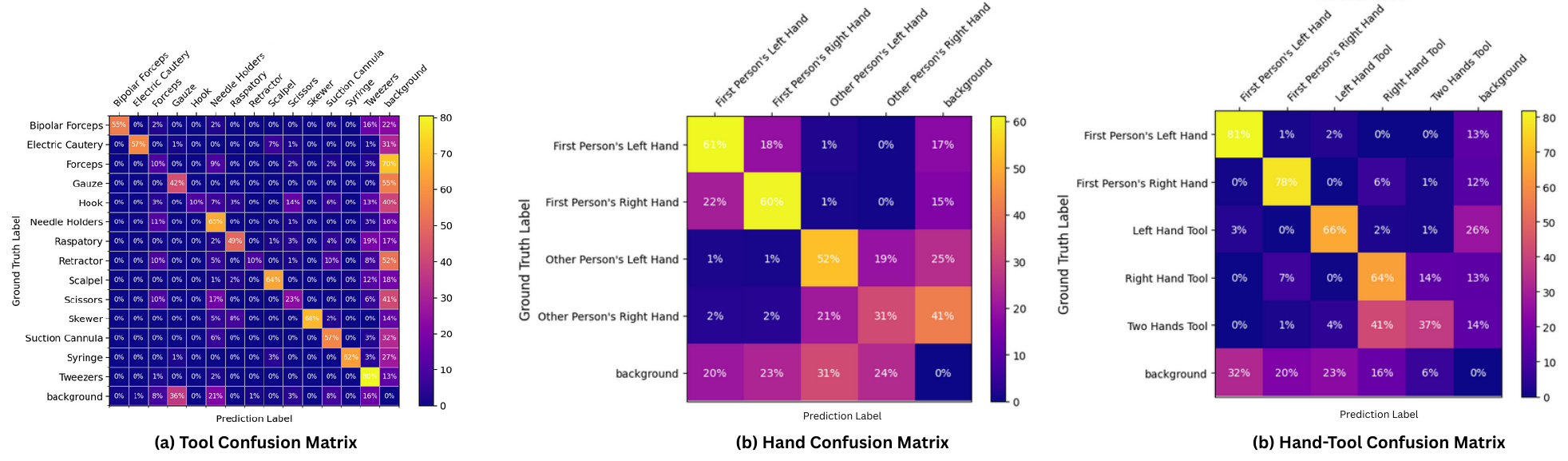}
\caption{Mask2Former Confusion Matrix over the three tasks. \textbf{(a)} Tool Confusion Matrix. \textbf{(b)} Hand Confusion Matrix \textbf{(c)} Hand-Tool Confusion Matrix}
\label{fig: confusion_matrix}
\end{figure*}

\noindent \textbf{Surgical Tool Segmentation}: For Surgical Tool detection, the box mAP are 36.7\%, 47.3\% and 39.2\% and the segmentation mAP are 29.1\%, 36.7\%, 40.9\% and 37.0\% for the respective Mask-RCNN, QueryInst, Mask2Former and SOLOv2 models. The experimental results are presented in table \ref{tab:benchmark}. QueryInst achieves the highest performance in terms of the mAP metric for surgical tool detection tasks with bounding box but mask2former outperforms the rest of the models in terms of mAP for per-pixel surgical tool segmentation. We observe some confusion in the models for tools with similar shapes in figure \ref{fig: confusion_matrix}. Moreover, each unique tool's average precision strongly depends on the tool's appearance frequency. The unbalance in tool's appearance offers a great disparity in mAP prediction for each tool. It leads to underperforming models induced by a lack of data on certain tools.  The overall results need to be improved, but similar to EgoSurgery-Tool results ~\cite{fujii2024egosurgeryphase,fujii2024egosurgerytool}, the results are encouraging for the future of practical tool segmentation in the context of open surgery. 

\noindent \textbf{Hand / Hand-Tool Segmentation}: The benchmark results presented in table \ref{tab:benchmark} for hand segmentation highlight great performances of Mask RCNN with a box mAP of 63.8\% and a mask mAP of 61.9\%. However, Mask R-CNN seems underperformant in surgical tool segmentation tasks compared to the three others. Mask2former consistently outperforms Mask-RCNN, QueryInst and SOLOv2 in segmentation mAP. In terms of bbox mAP for Hand-Tool detection, QueryInst achieves the best performance. Out of the four models overall, mask2former is superior if tool segmentation is part of the task. In the case of only hand segmentation, Mask R-CNN outperforms the other three models. Figure \ref{fig: confusion_matrix} reveals a higher degree of confusion among other hands, which are less distinct and less precisely defined compared to the owner's hands. Nevertheless, these tasks performances clearly show promising results for real applications in open surgical operations. 

\noindent \textbf{Comparison with Other Dataset}: We compare QueryInst Hand and Hand-Tool Segmentation performance on our testing set for different training data. Regarding Hand Segmentation, training on our dataset significantly outperforms training on EgoHands. Similarly, we observe superior results when training on our dataset as opposed to the Kitchen VISOR dataset. This strongly indicates a domain transfer issue from the representation of hands and tools in daily activities to open surgical ones. Our dataset serves as a valuable asset, enabling the learning of novel representations of surgical objects in the challenging environment of open surgery.

\begin{table*}[tb]
    \caption{Performance comparison of QueryInst trained on EgoSurgery-HTS and model trained on existing hand segmentation and hand-tool segmentation datasets.}
    \centering
    \resizebox{\textwidth}{!}{
        \begin{tabular}{cc}
            \begin{minipage}{0.49\textwidth}
                \centering
                (a) Hand segmentation\\
                \begin{tabular}{lcc}
                    \toprule
                    Dataset & mAP$^{\text{box}}$ & mAP$^{\text{mask}}$ \\
                    \midrule
                    EgoHands~\cite{zhang2022fine}  & 8.3 & 6.3 \\
                     \rowcolor{Gray} EgoSurgery-HTS & \textbf{54.0} & \textbf{50.7} \\
                    \bottomrule
                \end{tabular}
            \end{minipage}
            &
            \begin{minipage}{0.49\textwidth}
                \centering
                (b) Hand-tool segmentation\\
                \begin{tabular}{lcc}
                    \toprule
                    Dataset & mAP$^{\text{box}}$ & mAP$^{\text{mask}}$ \\
                    \midrule
                    VISOR-HOS~\cite{Darkhalil2022VISOR} & 13.0  & 11.4 \\
                     \rowcolor{Gray} EgoSurgery-HTS &  \textbf{55.2} & \textbf{49.4} \\
                    \bottomrule
                \end{tabular}
            \end{minipage}
        \end{tabular}
    }
    \label{fig:comparison_models}

\end{table*}

\subsection{Qualitative results}

\begin{figure*}[tb]
\centering
\includegraphics[width=\textwidth, trim= 0 10cm 0 0, clip]{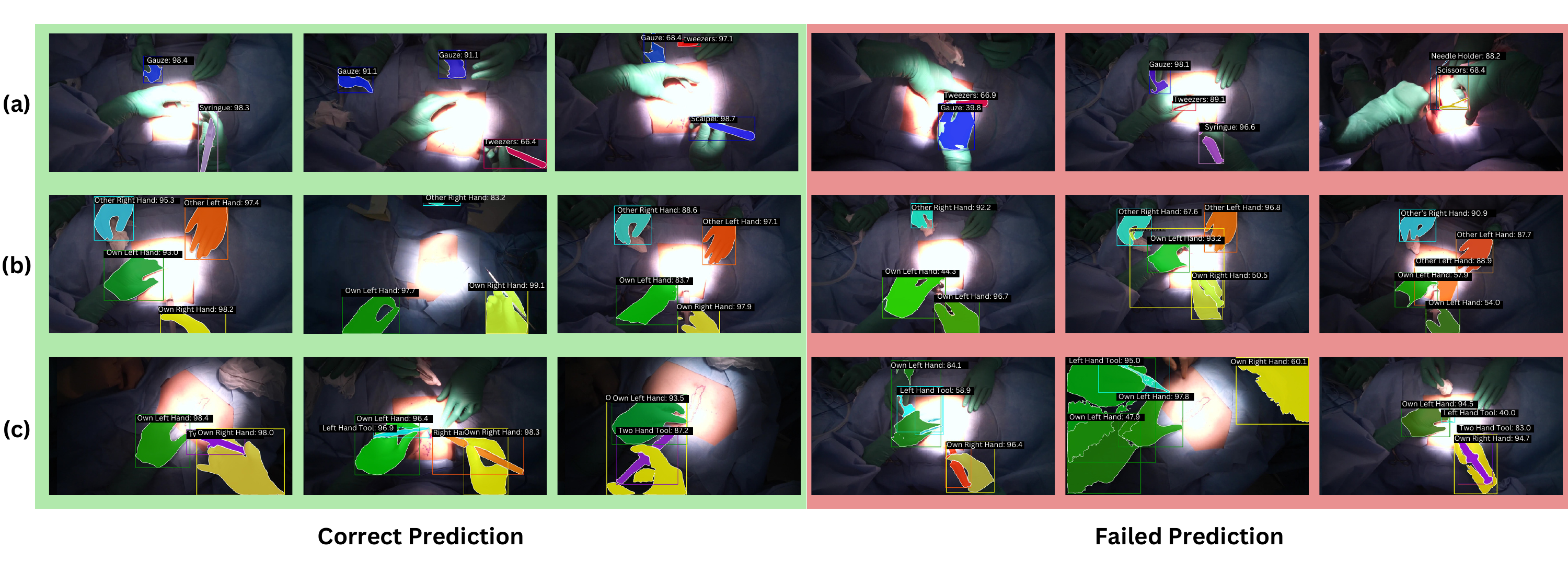}
\caption{Overview of the Mask2former model predictions on the three tasks. \textbf{(a)}Tool Segmentation task. \textbf{(b)}Hand Segmentation task. \textbf{(c)} Hand-Tool Segmenation Task.  }
\label{fig:modelpred}
\end{figure*}
\begin{figure*}[tb]
\centering
\includegraphics[width=\textwidth, trim= 0 10cm 0 1.75cm, clip]{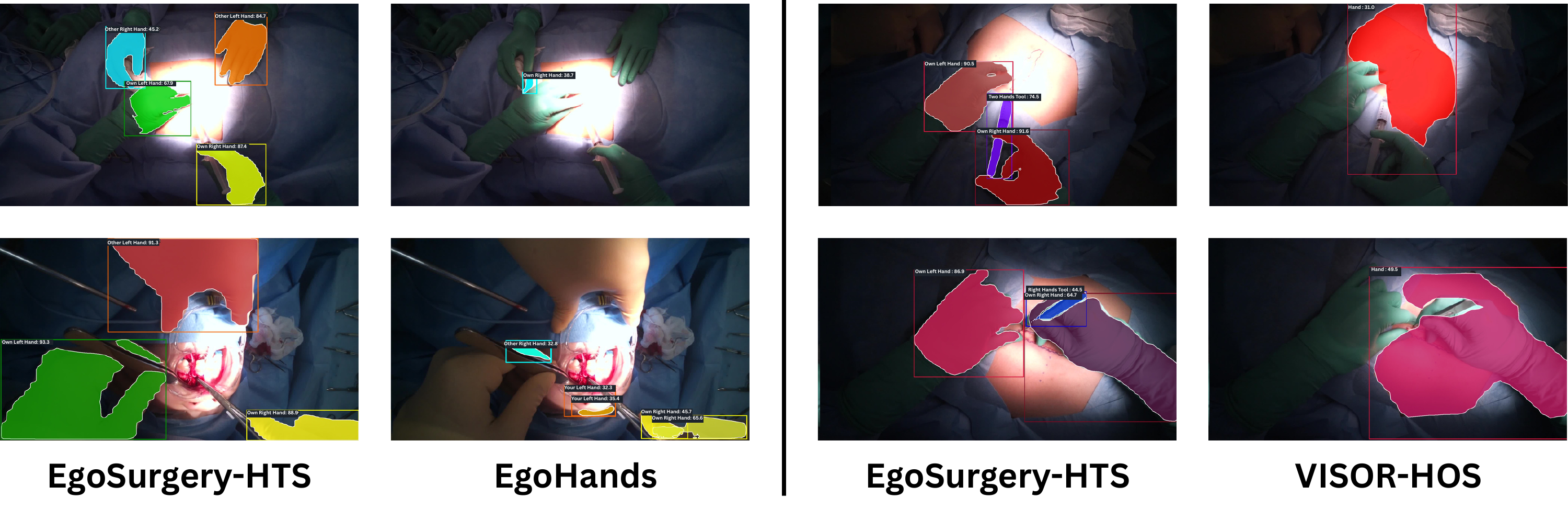}
\caption{Qualitative predictions of QueryInst trained on EgoSurgery-HTS dataset versus
trained on Egohands  \textbf{(left)} for  hand segmentation and  trained on VISOR-HOS  \textbf{(right)} for
hand-tool segmentation task  }
\label{fig:quantitative_comp}
\end{figure*}

The qualitative performance of the Mask2former model from the baseline is presented in the figure \ref{fig:modelpred}. The model successfully segmentates surgical tools, hands, and hand-tool accross many types of surgery. Nevertheless, we can observe omissions, errors, or imaginary tools predicted by the models in bad scenarios. This can be explained by the difficult distribution of luminosity, contrasts, shapes, and texture during open surgery video analysis. Nevertheless the models demonstrate robust performance even in suboptimal scenarios, indicating that they have successfully learned critical features essential for accurate hand-tool segmentation in the demanding conditions of open surgery.
Furthermore, the models demonstrate significantly improved performance when trained on our dataset compared to existing ones, particularly in the context of open surgery, as clearly evidenced by the results shown in figure \ref{fig:quantitative_comp}.

\section{Conclusion and Future Work}

In this paper, we introduce EgoSurgery-HTS, the first egocentric open-surgery segmentation dataset, including all hands and surgical tool instances plus the association of hand and tool. The dataset is composed of raw surgical videos of surgery with extensive segmentation annotations of every instance present. We define three tasks to improve the understanding of the egocentric open surgery scene through our dataset : Hand Segmentation, Tool Segmentation, and Hand-Tool Segmentation. We demonstrated the benefit of using EgoSurgery-HTS compared to other egocentric hand-tool datasets through pretrained model evaluations. A benchmark for each task is proposed as a reference for future evolution of the models evaluated on this dataset. 

Despite promising results, mAP results are still not acceptable for critical surgery applications. The continuing hurdles are to improve model for better segmentation to become usable during real case application. An auspicious next step is to focus on minimizing the challenges of segmenting heavily imbalanced data. Upcoming research efforts will concentrate on collecting more tool instances to equilibrate the tool distribution and increase the robustness of the model through additions of different  egocentric open surgery environment. 

\bibliographystyle{splncs04}
\bibliography{mybibliography}

\end{document}